\documentclass[letterpaper, 10 pt, conference]{ieeeconf}
\usepackage{times}
\usepackage{subfigure}
\usepackage{amsmath,amssymb,amsopn,amstext,amsfonts}
\usepackage{cancel}
\usepackage[space]{cite}
\usepackage{pdfsync}
\usepackage{balance}
\usepackage{color}
\usepackage{mathtools}
\usepackage{algorithm}
\usepackage{algorithmic}
\usepackage{bm}

\usepackage{diagbox}
\usepackage{float}
\usepackage{epstopdf}
\usepackage{pifont}
\usepackage{fixltx2e}
\usepackage{amsmath}
\usepackage{multirow}
\usepackage{url}
\usepackage{verbatim}
\usepackage[linkcolor=black,citecolor=black,urlcolor=black,colorlinks=true]{hyperref}
\usepackage{graphicx}
\usepackage{caption}
\usepackage{flushend}

\graphicspath{{figures/}}
\DeclareGraphicsExtensions{.png,.jpg,.eps,.pdf}
\IEEEoverridecommandlockouts
\title{\LARGE \bf Automatic Parameter Adaptation for Quadrotor Trajectory Planning}
\author{Xin~Zhou, Chao~Xu, and~Fei~Gao
\thanks{This work was supported by the National Natural Science Foundation of China under grant no. 62003299 and 62088101. All authors are with the College of Control Science and Engineering, Zhejiang University, Hangzhou, China, and Huzhou institute of Zhejiang University, Huzhou, China. (email: {\tt\small $\{$iszhouxin, cxu, fgaoaa$\}$@zju.edu.cn}).
 Corresponding author: Fei Gao.}}
\begin{document}

\maketitle
\thispagestyle{empty}
\pagestyle{empty}

\begin{abstract}
Online trajectory planners enable quadrotors to safely and smoothly navigate in unknown cluttered environments.
However, tuning parameters is challenging since modern planners have become too complex to mathematically model and predict their interaction with unstructured environments.
This work takes humans out of the loop by proposing a planner parameter adaptation framework that formulates objectives into two complementary categories and optimizes them asynchronously.
Objectives evaluated with and without trajectory execution are optimized using Bayesian Optimization (BayesOpt) and Particle Swarm Optimization (PSO), respectively. 
By combining two kinds of objectives, the total convergence rate of the black-box optimization is accelerated while the dimension of optimized parameters can be increased.
Benchmark comparisons demonstrate its superior performance over other strategies.
Tests with changing obstacle densities validate its real-time environment adaption, which is difficult for prior manual tuning.
Real-world flights with different drone platforms, environments, and planners show the proposed framework's scalability and effectiveness.

\end{abstract}

\section{Introduction}
\label{sec:introduction}

Tremendous advances in quadrotor trajectory planning have brought aerial robots into various complex wild environments with different hardware specifics. 
To be adaptable in diverse applications, planners should adjust the parameters accordingly, where domain knowledge is always necessary.
This requirement limits the broad application of the planning algorithms to general users.  
Furthermore, as the codebase becomes more and more complex, parameters could be highly coupled, leading even expert to get confused and only adjust around the default setting, which is tuned for other cases. 
The potential of the planner is still not well exploited.

For example, desired flight speed, a widely used and intuitive parameter relevant to flight time and control error, is always hard to determine when users face a new environment or a new drone. 
To address this problem, some works propose specific handcrafted rules\cite{quan2021eva} or learning-based methods \cite{richter2014high} for speed altering.
However, it is impractical to design a separate system for every parameter, even more so for abstract ones.
Although some robot navigation tutorials \cite{zheng2021ros} are kindly provided, they only focus on specific planners used for vehicles on 2-D surfaces. 
There is still a lack of insight into automatic parameter adaption for online aerial robot trajectory planning.

\begin{figure}[t]
	\centering
	\includegraphics[width=1.0\linewidth]{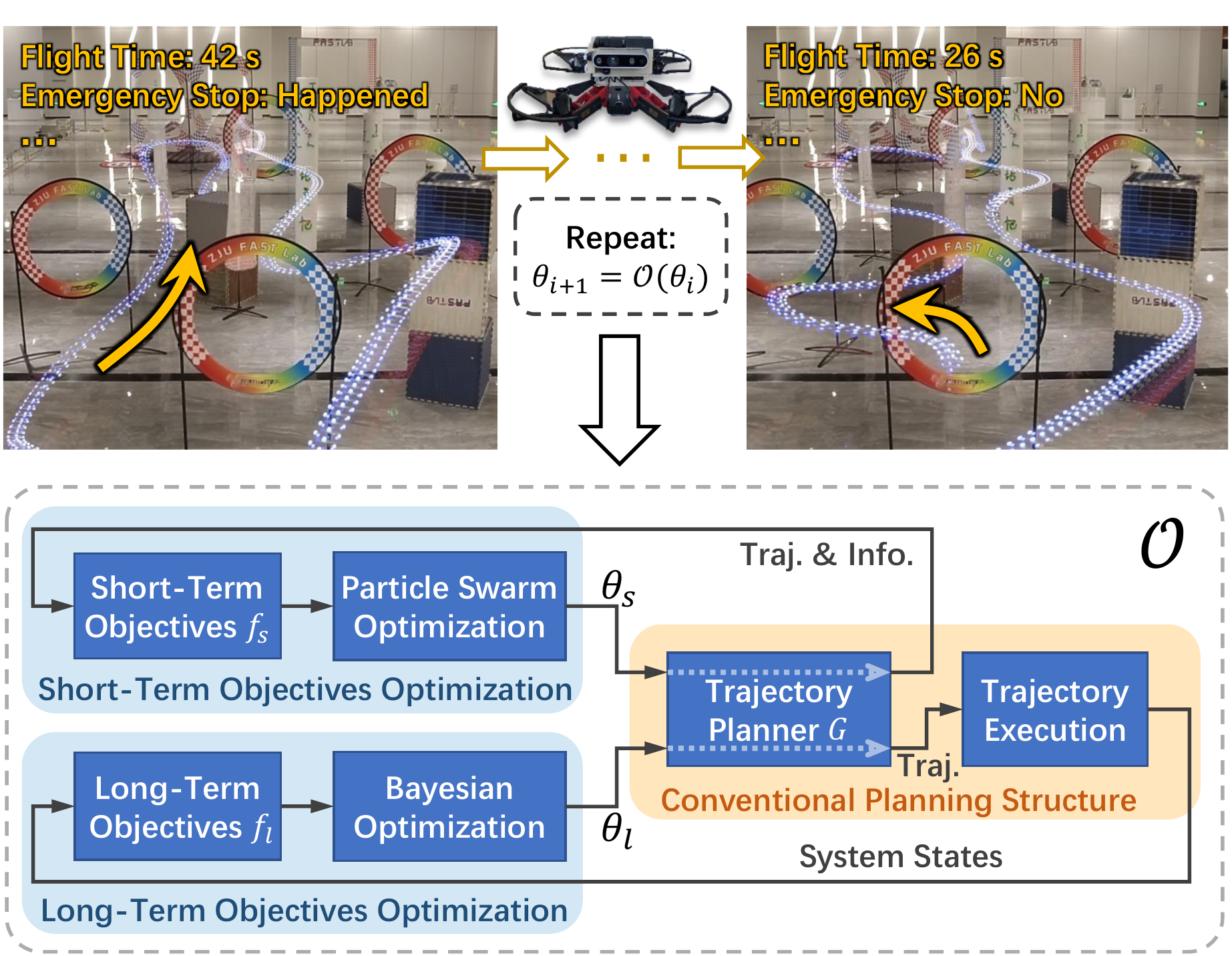}
	\caption{ The proposed complementary parameter adaption framework $\mathcal{O}$. $\theta_s$ and $\theta_l$ are parameters optimized for short-term and long-term objectives, respectively. \textit{Traj.} is short for \textit{Trajectory} and \textit{Info.} represents \textit{Information}. The loop for short-term objectives optimization runs at a significantly higher frequency than the other loop.}
	\label{pic:system}	
	\vspace{-0.8cm}
\end{figure}

In this work, we propose a general and effective parameter tuning framework. 
This framework divides the desired performance objectives into \textit{short-} and \textit{long-term} categories.
Short-term objectives like computing time can be found immediately after generating a new trajectory, while long-term objectives like tracking error require a finite period of trajectory execution.
Leveraging a common feature of online trajectory planners that they should compute and evaluate a new trajectory within a limited time to enforce real-time reaction to sensed obstacles, short-term objectives can be repeatedly evaluated and optimized at a high frequency for up to several hundreds of times per second, even in parallel.
Therefore a low-complexity, sampling-based optimizer is suitable.
By contract, evaluating long-term objectives is expensive; therefore, data-efficient algorithms always demand higher computation, but converges with fewer iterations are preferred.
The specific optimizers in the proposed framework are not restricted.
As a validated option, in most parts of this work, we adopt \textit{Particle Swarm Optimization (PSO)} and \textit{Bayesian Optimization (BayesOpt)} \cite{frazier2018tutorial} for short-term and long-term objective optimization, respectively.
Utilizing their properties produces a complementary parameter adaption framework, as shown in Fig. \ref{pic:system}.

Benchmark comparisons validate that our performance exceeds that of using the default parameters or a single optimizer. 
Furthermore, the experiment shows that the proposed framework successfully found satisfactory parameters for two widely used quadrotor trajectory planners \cite{zhou2021raptor, zhou2020ego} with different parameters even though they run on different quadrotor platforms in different environments.
The contributions of this paper are:
\begin{enumerate}
	\item 
	We propose a general framework that optimizes various short-term and long-term objectives by tuning parameters of quadrotor trajectory planners without human assistance.
	\item
	We integrate the proposed method into a fully autonomous quadrotor system and validate its effectiveness, and will release our software \footnote{https://github.com/ZJU-FAST-Lab/Auto-Param-Adaption} for the reference of the community.
\end{enumerate}

\section{Related Work}
\label{sec:related_work}

Tuning the parameters of an existing trajectory planner \cite{xiao2022appl, xiao2020appld, wang2021appli, wang2021apple, xu2021applr, bhardwaj2020differentiable, teso2019predictive} rather than optimizing the trajectory directly from a black box \cite{ryou2021multi, loquercio2021learning} can take full advantage of modern complex but high-performance planners. 
Among them, some approaches \cite{xiao2020appld, wang2021appli, wang2021apple} get inspiration from human assistance, while others take the humans out of the loop \cite{xu2021applr, bhardwaj2020differentiable, teso2019predictive} using neural networks to learn the parameter tuning policies from environments.

This paper focuses on automatic parameter adaption for online quadrotor trajectory planners without human assistance. 
A class of traditional approaches belonging to the discipline of control theory construct the performance metric as a differentiable function over tuned parameters and optimize using optimal control or gradient-based methods \cite{grimble1984implicit, trimpe2014self}.
However, a mathematical system model in trajectory planning is always intractable to construct, owning to complex internal mechanisms and unpredictable real-world noise and disturbance \cite{loquercio2022autotune}. 
Instead of formulating the objective function analytically, some other methods use surrogate models such as Gaussian Process (GP) to approximate that function \cite{marco2016automatic, ryou2021multi}. 
The accuracy of the estimation increases with the number of observations, and the most promising parameter set is acquired from the latest GP using some criteria called acquisition functions. 
This widely used algorithm is Bayesian optimization \cite{frazier2018tutorial}.
If there are no assumptions about the model, some sampling-based, model-free strategies such as random search can be widely adopted into, for example, hyper-parameter tuning in machine learning \cite{bergstra2012random}. 
Extra legend improvements rooted in purely random search considering historical information have evolved into, for example, simulated annealing \cite{van1987simulated} and particle swarm optimization \cite{kennedy1995particle}.
In aerial flight, Loquercio et al. \cite{loquercio2022autotune} use Metropolis-Hastings sampling belonging to the model-free category to adjust controller parameters for agile flight.
Both \textit{surrogate-model-based} and \textit{model-free} strategies belong to the category of derivative-free optimization.

\section{Methodology and Discussion}
\label{sec:methodology}

\subsection{Problem Definition}
\label{sec:Problem_Definition}

We consider a given quadrotor trajectory planning algorithm that may contain several components, including but not limited to a finite state machine, path search methods like A*, local trajectory optimization, and some safety guarantee mechanisms like emergency stop.
The planner $G:\mathcal{S}\times \Theta \rightarrow \mathcal{L}$ takes system state $s \in \mathcal{S}$ under the parameter set $\theta \in \Theta \subset \mathbb{R}^d$ as inputs and produces a time-parameterized trajectory $l = p(t) \in \mathcal{L}$ in 3-D space of probability $P(l_{i+1}|s_{i}, \theta_{i})$, where $i$ is the iteration index.
Time $t \in [0, T_m]$, the time domain of the trajectory.
System state space $\mathcal{S}$ contains obstacle maps, the current and the target drone state, and the currently executed trajectory.
The objective function $f(s(\theta_{i}),l(\theta_{i}))$ evaluated from $\mathcal{S}$ and $\mathcal{L}$ is computed over a finite time horizon with randomness and system noise.
Then optimizing $f(\theta)$ becomes a typical multi-armed bandit problem, and the system is a standard Markov Decision Process (MDP) of the tuple ($\mathcal{S}, \Theta, P, f$), where the objective $f:\Theta \rightarrow \mathbb{R}$, but the functional dependence of $f$ on $\theta$ is unknown.
The overall parameter tuning problem is 
\begin{equation}
	\theta^* = \min_{\theta \in \Theta} \mathbb{E}_{\{l | s, \theta\}} \left[ f(\theta) \right],
\end{equation} 
where the notion $\mathbb{E}$ indicates expectation.

The basic strategy is iteratively selecting parameters $\theta_{i+1}=\mathcal{O}(\theta_{i})$ based on the black-box optimization strategy $\mathcal{O}$ and evaluating the corresponding function values $f(\theta)$ until the termination criteria are met.

\subsection{The Complementary Framework}
As introduced in Sec. \ref{sec:related_work}, quadrotor trajectory planning interacting with non-convex, complex environments, and containing various user-defined state machines is a nondifferentiable system. 
Therefore, derivative-free optimizations are relatively suitable for parameter tuning for trajectory planning.
Between the surrogate-model-based and model-free approaches, the former always shows faster convergence but at the cost of higher computation to construct the surrogate model and determine which parameter set $\theta_{i+1}$ to choose next. 
This feature is particularly significant in its representative method --- Bayesian optimization, which is thus more suitable for expensive-to-evaluate objective functions \cite{frazier2018tutorial}.
By contrast, model-free strategies always converge slower but calculate less.

Investigating state-of-the-art, online trajectory planners for quadrotors \cite{zhou2019robust, tordesillas2019faster, zhou2020ego}, the substantial differences of how objectives are observed divide objectives into two categories, as mentioned in Sec. \ref{sec:introduction}.
In one category, objective values can be determined as soon as a trajectory is generated, like computing time, smoothness, spacing to obstacles, etc.
In contrast, those of the other category are observed only after executing the trajectory, like tracking error, emergence cases encountered, time to complete the mission, etc.
In this work, we name them \textit{short-term} objectives $f_s(\theta_s)$ and \textit{long-term} objectives $f_l(\theta_l)$, respectively, where $\theta_s \in \Theta_s  \subseteq \Theta$, $\theta_l \in \Theta_l  \subseteq \Theta$.
Since the computation time of a trajectory is usually several orders of magnitude less than the execution time (milliseconds order to seconds order) for online planners to enforce real-time, short-term objectives can be evaluated at a significantly higher frequency than long-term objectives.
This difference is further amplified in multithreading.
Therefore it is always effortless to get a bunch of observations for short-term objectives.
By contrast, long-term objectives are relatively expensive to evaluate.  

Utilizing the above features of optimization methods and online planners, it is natural to adopt model-free methods for short-term objectives optimization and surrogate-model-based methods for long-term objectives.
Combining the above two optimizations as in Fig. \ref{pic:system} produces a complementary automatic parameter adaption framework $\mathcal{O}$ where the short-term optimization loop explores the solution space extensively, and the long-term optimization loop considers more information when executing the trajectory.
Both loops provide extra knowledge to each other therefore improving the total convergence.
The parameters are tuned in low-fidelity simulation at the first phase and then moved to a real drone for high-fidelity fine-tuning with narrower parameter boundaries for safety at the second phase.

\subsection{Discussion of the Adopted Optimizers}

In this work, PSO and BayesOpt are adopted for optimizing short-term and long-term objectives, respectively.
However, any derivative-free optimization methods can be appropriate according to the problem specifics.
For example, in the experiment of Fig. \ref{pic:obstacle_distribution}, BayesOpt is used for optimizing $f_s$ since the number of evaluations is limited.

\subsubsection{Particle Swarm Optimization}
Particle Swarm Optimization belonging to model-free, random-sampling-based approaches has shown effectiveness in automatic parameter tuning in other scenarios\cite{solihin2011tuning, cai2013improved}. 
This method maintains a population of $N_p$ candidate solutions $\theta_{n,i}, n \leq N_p$, dubbed particles, at iteration $i$, and each solution updates independently according to the simple mathematical formula
\begin{align}
	& v_{n,i+1} = w v_{n,i} + \phi_x r_x (p_n - \theta_{n,i}) + \phi_g r_g (g - \theta_{n,i}), \\
	& \theta_{n,i+1} = \theta_{n,i} + v_{n,i+1}
\end{align}
where $w, r_x, r_g$ are three user-defined hyper-parameters balancing the weight of three terms, $\phi_x, \phi_g \sim U(0,1)$ are uniformly distributed random numbers, $p_n$ is the current best of the $n$-th particle while $g$ is the best of all particles.
When $i=0$, $v_{n,0}$ and $\theta_{n,0}$ are generated randomly within the domain.
As PSO only stores the current best, the time complexity is constant over the sampling number.

\subsubsection{Bayesian Optimization}
Bayesian optimization comprises two parts, i.e., a surrogate model to estimate the objective function and an acquisition function to decide which point to evaluate next.
For the first part, \textit{Gaussian Process} (GP) model is typically adopted which can quantify the uncertainty.
GP estimates the posterior distribution of function value $f(\theta) | f(\theta_{1:i})$ according to the prior $f_{prior}(\theta) \sim \mathcal{N}(\mu_0(\theta), \sigma_0^2(\theta))$ and previous $i$ evaluations $f(\theta_{1:i}) \sim \mathcal{N}(\mu(\theta_{1:i}), \Sigma(\theta_{1:i}, \theta_{1:i})) \in \mathbb{R}^i$ following
\begin{equation}
	\label{equ:GP}
	\begin{split}
	&f(\theta) | f(\theta_{1:i}) \sim \mathcal{N} (\mu(\theta), \sigma^2(\theta)),\\
		&\mu(\theta) =\Sigma(\theta, \theta_{1:i}) \Sigma(\theta_{1:i}, \theta_{1:i})^{-1}(f(\theta_{1:i})-\mu(\theta_{1:i})) + \mu_0(\theta), \\
		&\sigma^2(\theta) = \sigma^2_0(\theta) - \Sigma(\theta, \theta_{1:i}) \Sigma(\theta_{1:i}, \theta_{1:i})^{-1}\Sigma(\theta_{1:i}, \theta),
	\end{split}
\end{equation}
where the covariance matrix $\Sigma$ is computed by a user-defined kernel function $\mathcal{K}:\mathbb{R}^d \times \mathbb{R}^d \rightarrow \mathbb{R}$ describing the correspondence between two points in $d$-dimensional space.
The higher the outputs of $\mathcal{K}$, the higher the correspondence.

Then various acquisition functions considering uncertainties can be used, such as expected improvement \cite{jones1998efficient} or entropy search \cite{hennig2012entropy}.
This work adopts the former one that optimizes
\begin{equation}
	\theta_{i+1} = \max_{\theta \in \Theta} \mathbb{E}\left[\max(f_{1:i}^+ - f_{GP}(\theta), 0)\right],
\end{equation}
where $f_{1:i}^+=\min(f(\theta_{1:i}))$, $f_{GP}(\theta) = f(\theta) | f(\theta_{1:i})$.

BayesOpt considers all previous observations, thus producing more promising solutions than PSO with limited samples. 
However, the time complexity over sample number $N$ is $O(N^3)$, owing to \textit{Cholesky Decomposition} for the matrix inversion in Eq. \ref{equ:GP}.
Therefore, BayesOpt is more suitable for expensive-to-evaluate objective functions.

\subsubsection{Tolerance to Noisy Observations}

As discussed in Sec. \ref{sec:Problem_Definition}, evaluations of $f(\theta)$ contains randomness. 
In general, researchers sample for multiple replications and compute the average to reduce the uncertainty as discussed in the Sec. 1.2.3 of \cite{chen2015stochastic}.
The number of replications can be determined using probabilistic characterization of the variability \cite{chen2015stochastic}.
Fortunately, PSO shows tolerance to noisy observations\cite{parsopoulos2001particle}.
Also discussed in \cite{parsopoulos2001particle} that, in many cases, the presence of noise seems to help PSO to avoid local minima of the objective function and locate the global one. 

BayesOpt using GP to estimate the system model can naturally deal with uncertainty.
In GP, we assume zero-mean, Gaussian-distributed observation noise $\omega \sim \mathcal{N}(\mathbf{0}_{i\times 1}, \rm{diag}(\sigma^2_{\omega}(\theta_{1:i})))$ with $\theta_{1:i}$ denotes $i$ independent sampled points.
This noise only affects the covariance matrix of sampled points
\begin{equation}
	\begin{split}
		\Sigma(\theta_{1:i}, \theta_{1:i}) = & \mathcal{K}(\theta_x, \theta_y) | _{x, y \in \{1,2,\cdots,i\}}  \\ 
		&+ \rm{diag}(\sigma^2_{\omega}(\theta_{1:i})).
	\end{split}
\end{equation}

However, as GP takes all previous observations for function approximation, the GP model becomes inaccurate when the environment, such as obstacle density, changes.
In order to relieve this effect, methods such as \textit{Online Context Prediction} \cite{xiao2020appld} can be used to automatically match different parameter sets from a library for different scenarios.

Another notable problem is that, as $\theta_s$ is part of the changing environments from the perspective of BayesOpt, 
previous $f_l(\theta_l)$ evaluations based on suboptimal $\theta_s$ become less convincing when PSO produces a new $\theta_s$.
Therefore, we reduce the fidelity of outdated $f_l(\theta_l)$ observations by enlarging their initial variance $\sigma_0^2(f_l(\theta_l))$ by a discount factor $\lambda_d \in (0,1)$
\begin{equation}
	\sigma^2(f_l(\theta_l)) = \frac{\sigma_0^2(f_l(\theta_l))}{\lambda_d^{n_d}},
\end{equation}
where $n_d$ is the times of best $\theta_s$ change since the corresponding $f_l(\theta_l)$ has been evaluated.

In practice, if there are no initial available parameters provided, short-term objectives $\theta_s$ are optimized prior to the long-term part to find a relatively reasonable $\theta_s$.
Only after that will the long-term optimization start.

\subsection{General-Purpose Objective Functions}

\subsubsection{Short-Term Objectives}

Theoretically, any objectives that are relevant to parameters $\theta_s$ can be optimized.
In practice, to achieve overall performance, some common objectives are recommended, especially for short-term objective functions $f_s$ which directly reflect the trajectory optimization quality inside the planner $G$, and thus constructing appropriate $f_s$ is fundamental.
The planner $G$ has already optimized some widely used objectives, such as trajectory smoothness, dynamical feasibility, and collision-free to obstacles.
Therefore they are generally not incorporated into $f_s$.
Instead, we consider (1) computing time $t_c$ to compute a trajectory, and (2) the number of failed trials $n_s$ before successfully generating a trajectory (which is also considered as a success rate metric).
Note that $t_c$ and $n_s$ can not be incorporated into our traditional gradient-based planning framework \cite{zhou2020ego}, as the correspondence between these noisy objectives and decision variables can not be mathematically formulated.

Investigating the requirements of quadrotor online trajectory planning, which has to timely react to the latest environments mapped using onboard sensors, we have to bound the computing time.
Also, a shorter computing time is still desired.
Therefore we formulate the computing time penalty $J_t$ as 
\begin{equation}
	J_t = t_c + w_{\tau} \max (0,t_c - \tau_t)^2,
\end{equation}
where $\tau_t$ is the real-time threshold and $w_{\tau} \gg 1$ to enforce the real-time constraint.

$n_s$ is considered under the phenomenon that most modern online planners for quadrotors are robust to intermittent failures \cite{zhou2019robust,usenko2017real,tordesillas2019faster,zhou2020ego}. 
When a planning attempt fails, the quadrotor keeps executing the previous trajectory, and the planner will quickly perform another trial.
However, executing a previous trajectory is not safe because it only guarantees safety to previously mapped obstacles, and thus, less trial is still desired.
Therefore the penalty of failures $J_s$ is defined as $J_s = n_s$.

Then $f_s$ is defined as a weighted combination
\begin{equation}
	\label{equ:f_s}
	f_s = w_t J_t + w_s J_s,
\end{equation}
where $w_t, w_s \geq 0$ are weights for balancing two objectives.

\subsubsection{Long-Term Objectives}

Optimizing well-defined, short-term objectives builds a high-performance planner in static states.
Then we can consider more factors by incorporating real-time flight metrics on multiple consecutive trajectories to formulate the long-term objectives $f_l$.
In general, $f_l$ is more direct about users' expectations of a trajectory planner, such as (1) shorter flight time $t_f$, (2) minor trajectory tracking error $e_t$, (3) fewer emergency cases encountered $n_e$, and (4) zero collision cases $n_c$, as have been mentioned in Sec. \ref{sec:introduction}.

Reducing flight time is desired in scenarios like search and rescue, and transportation. 
In our work, we command the drone to fly through a series of given waypoints and then return to the start point. 
The time spent from start to return is recorded as $t_f$, which constitutes the flight time penalty $J_f=t_f$.
Then BayesOpt process produces the next best candidate $\theta_l$, which is then used to perform another lap immediately.

Tracking error is the deviation of the drone position $p_d$ to desired trajectory position $p(t)$ at time $t$.
This deviation may result in a collision even though the executed trajectory is collision-free.
Traditional methods always improve tracking accuracy by designing better controllers \cite{lee2017trajectory}, while recent works \cite{ryou2021multi,foehn2021time} validate that the trajectory itself also matters.
Since the planner has already well-considered dynamical feasibility, tracking error becomes significant mainly in some particular cases, such as a fast and sharp turn, which may exceed the dynamical capability \cite{ryou2021multi}. 
Therefore we only record the maximum tracking error that occurred during a given time or a distance window.
The tracking error penalty $J_p$ is then defined as
\begin{equation}
	\label{eq:tracking_error}
	J_p = \max{(\|p_d - p(t)\|)}, 
\end{equation}
where $t \in [t_s, t_e]$, a predefined time domain.

Modern online trajectory planners are always equipped with emergency backup plans.
For example, in \cite{tordesillas2019faster}, the planner plans aggressive and conservative trajectories simultaneously.
When an emergency happens, the drone will immediately switch to the conservative one.
In \cite{zhou2020ego} and \cite{zhou2019robust}, an emergency stop will be triggered when a collision is forecast to happen within a time threshold $\tau_c$ while the planner is failed to find any acceptable trajectory to escape this situation.
As triggering an emergency action harms flight quality, we penalize the times of emergency actions $J_e = n_e$.

The times of collision to obstacles are also severely penalized by $J_c = n_c$.
Typically the weight accompanied to $J_c$ is far larger than the weight of other penalties.

Then the long-term objective function $f_l$ can be defined as weighted single objectives or any other reasonable forms.
One thing the developers should remember is that $f_s$ and $f_l$ should share consistent tendency. 
For example, both of them tend to fly aggressively or smoothly. 

\subsection{Transformation from Simulation to the Real World}
\label{sec:sim_to_real}

As poor parameters may lead to unsafe actions, the parameter optimization starts from simulation with similar obstacle density and drone specifics to the deployment scenarios.
After acquiring the optimized $\theta$, we set a narrower parameter boundary $\Theta|_{\theta_{low} \leq \theta \leq \theta_{up}}$ to perform high-fidelity real-word refinement safely.
This boundary moves around the current best parameters $\theta^*_i$ with a damping factor $\lambda_b \in (0,1)$, 
\begin{align}
	\theta_{up,i+1} &= \lambda_b \theta_{up,i} + (1-\lambda_b) (\theta^*_{1:i}+r_b), \\
	\theta_{low,i+1} &= \lambda_b \theta_{low,i} + (1-\lambda_b) (\theta^*_{1:i}-r_b),
\end{align}
where $r_b$ is half the width of the boundary, which balances safety and convergence rate.
If the optimization starts from a feasible solution already, the simulation phase can be skipped.
Other methods such as \textit{SafeOpt} \cite{sui2015safe} considering safety risk for Bayesian Optimization can also be adopted.

\section{Results}

In this section, we evaluate the characteristics of the proposed framework in the following aspects: 1) Benchmark comparisons with the default parameters, purely random sampling, PSO only, and BayesOpt only; 2) Effectiveness to changing and consistent environments; 3) Effectiveness to different quadrotors and trajectory planners.
The adopted online trajectory planner is our previously proposed EGO-Planner \cite{zhou2020ego}, with FAST-Planner \cite{zhou2019robust} added in Sec. \ref{sec:Real_word_Deployment}.
An open-source BayesOpt library MOE \cite{moe} is adopted with Square Exponential GP kernel $\mathcal{K}(\mathbf{x}_1, \mathbf{x}_2)=\alpha \exp (1/2 \times (\mathbf{x}_1-\mathbf{x}_2)^\mathrm{T}L(\mathbf{x}_1-\mathbf{x}_2))$ and the \textit{Expected Improvement} \cite{jones1998efficient} acquisition function used.
The simulations run on Intel core i7 9700K CPU at 5.0 GHz with 32 GB memory.

\subsection{Benchmark Comparisons}
\label{sec:Benchmark_Comparisons}

In the following tests, the drone is commanded to fly to a target 24-meters ahead and return.
The basic principle is to make the drone finish the task as quickly as possible; thus the short-term objective of Eq. \ref{equ:f_s} is adopted, and the long-term objective $f_l$ is defined as
\begin{equation}
	\label{eq:Benchmark_Comparisons}
	f_l = J_f + J_f J_c + w_e J_e = t_f+ t_f n_c + w_e n_e,
\end{equation}
where $w_e$ is the weight of the emergency stop penalty.
This definition penalizes the flight time $t_f$ directly and gives collision times $n_c$ a dominant cost to improve safety.
Note that we take the flight time $t_f$ as the collision weight directly because accidental collisions in agile maneuvers are always unavoidable, yet if a collision happens in smooth and slow flight, the parameters are seen as far from optimal.

In this benchmark, $\theta_s = [w_{col}, w_{dyn}, n_{pol}]$, where $w_{col}$ is \textit{collision avoidance weight}, $w_{dyn}$ is \textit{dynamical feasibility weight}, and $n_{pol}$ is \textit{the number of polynomial pieces}.
They are closely relevant to the trajectory optimization procedure in $G$.
$\theta_f = [v_{des}, h_{pl}, t_{pl}]$, where $v_{des}$ is \textit{desired velocity}, $h_{pl}$ is \textit{local planning horizon}, and $t_{pl}$ is \textit{re-planing interval}.
Other parameters like obstacle inflation and emergency stop threshold with clear safety boundaries are not optimized.

\begin{figure}[t]
	\centering
	\includegraphics[width=1.0\linewidth]{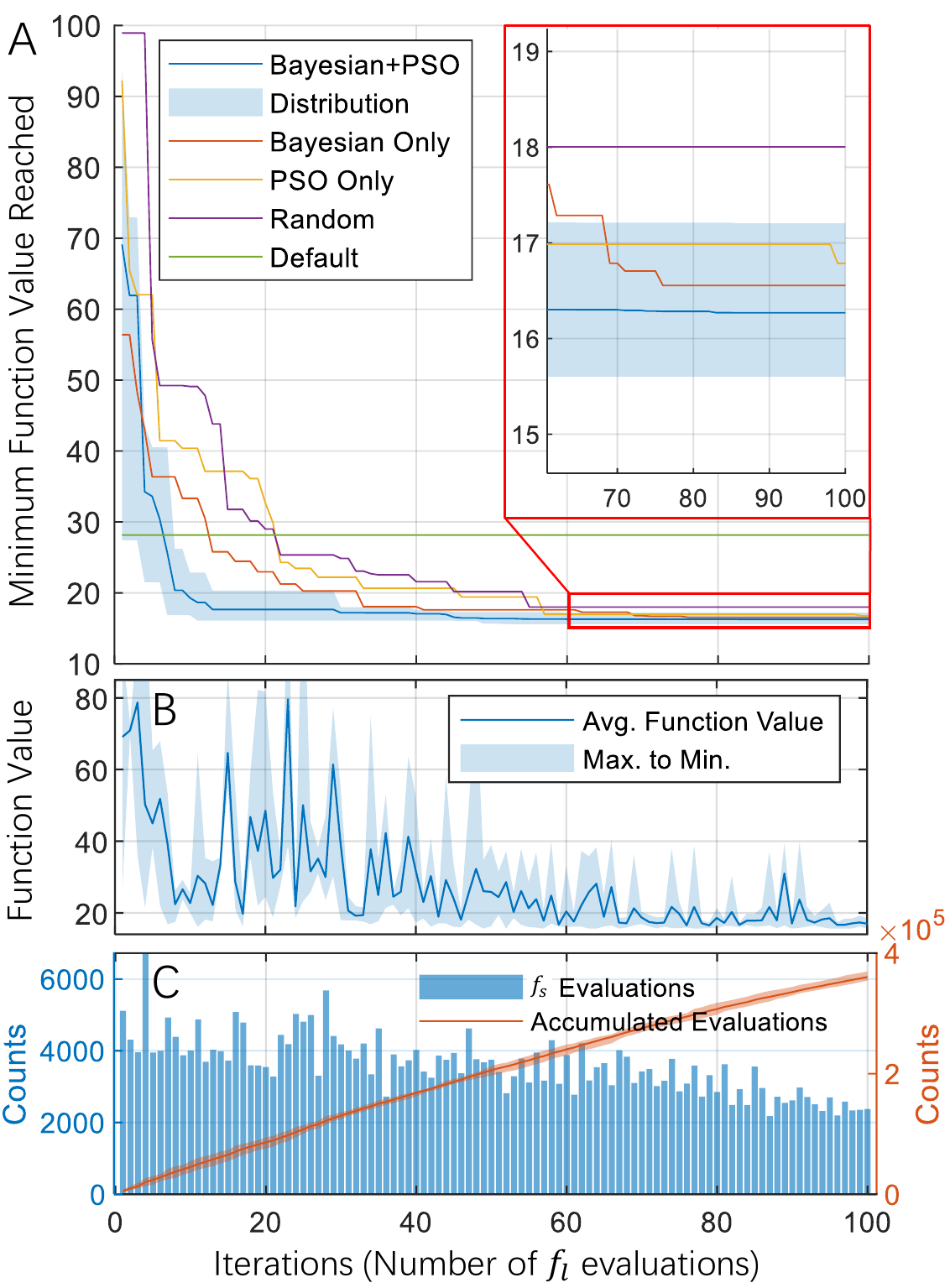}
	\caption{ Benchmark comparisons. (A) The curves record the latest minimum objective values acquired during the entire optimization. Solid curves are the average value, and the shading is the upper and lower bounds among five independent tests. Only the distribution for the \textit{Bayesian+PSO} is shown for figure clarity. The green curve added as a baseline is the average of five flight rounds using the default parameters. The upper-right figure is a magnified view of the final convergence. (B) The average $f_l$ value with low and up boundaries in each iteration. (C) The number of $f_s$ evaluations.}
	\label{pic:benchmark}	
	\vspace{-0.3cm}
\end{figure}

Except for the default parameter group, which does not require iterations, all the other tests iterate for 100 rounds.
Each optimizer as well as the default parameter group repeats five times in random maps with identical obstacle density.
The results are shown in Fig. \ref{pic:benchmark}.

In the first few iterations, the proposed framework converges significantly faster than other alternatives.
According to Eq. \ref{eq:Benchmark_Comparisons}, a large $f_l$ always indicates collisions encountered by all the methods at first. 
However, the proposed framework quickly managed to explore the safe spaces.
The notable thing is that splitting the objective $f$ into $f_s$ and $f_l$ allows evaluating $f_s$ with massive times.
As shown in Fig. \ref{pic:benchmark}B, during 100 $f_l$ evaluations, $f_s$ is called for about $4 \times 10^5$ times.
After each long-term evaluation, short-term parameters have been optimized for about 3500 iterations in the background. That is an implicit reason making long-term parameters converge faster in the \textit{Bayesian+PSO} method.
Except that, dividing $\theta$ into two parts significantly reduces the dimension of the parameter space for each optimizer, thus making the searching more efficient.
In other words, the number of parameters that can be simultaneously optimized is increased.

After 100 $f_l$ iterations, all the parameter tuning methods achieve superior performance than the default parameters.
The difference of the final results of three methods, i.e., \textit{Bayesian+PSO}, \textit{Bayesian Only}, and \textit{PSO Only}, are statistically insignificant, but they all beat \textit{Random Search} by about 10\%.

\subsection{Effectiveness to Changing and Fixed Environments}

\subsubsection{Changing Environments}
\begin{figure}[t]
	\centering
	\includegraphics[width=1.0\linewidth]{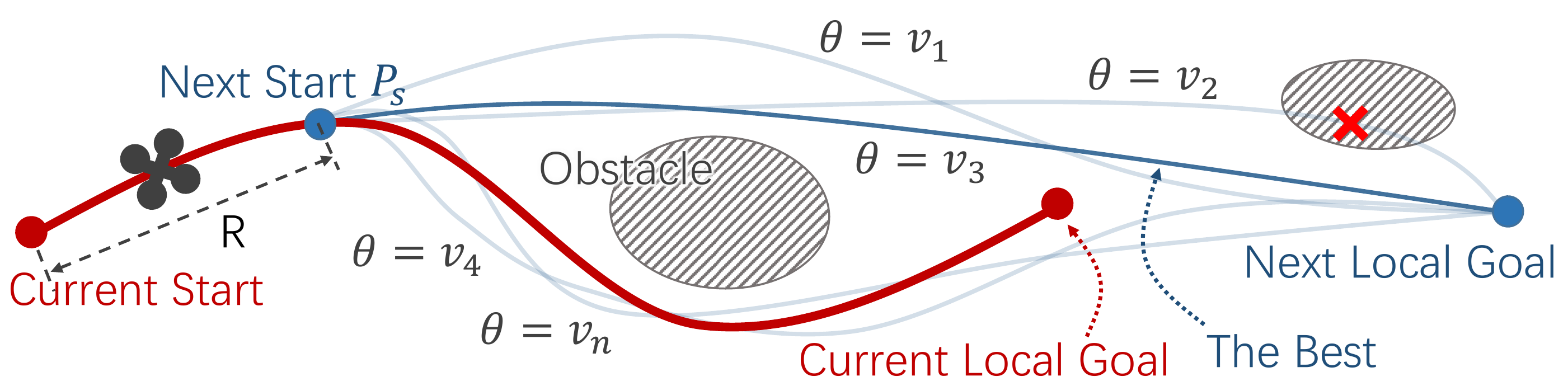}
	\caption{ Problem formulation for parameter tuning in changing environments. A bunch of trajectory planning start from a future point $P_s$ with different parameters $\theta$, then the one with the smallest objective function value will be selected to executed next.}
	\label{pic:changing_env_method}	
		\vspace{-0.8cm}
\end{figure}

Unlike previous works that take sensor readings as policy inputs \cite{richter2014high,xu2021applr}, we optimize parameters only based on the evaluation to planned trajectories.
Investigating the difference between sparse and dense environments, we find that the generated trajectories in dense places are more jerky than those in sparse scenarios, resulting in more significant tracking error.
Therefore, we directly penalize trajectory smoothness (time-integral of squared jerk) for adapting to changing obstacle density.
Further more, for shorter flight time and safety, we also penalize the trajectory executing time $t_{exe}$ and the smallest clearance $c_{obs}$ to obstacles.
Above objectives can be calculated without a real flight; therefore, they formulate a short-term objective function
\begin{align}
	\label{eq:changing_env}
	& f_s = w_t t_{exe} + w_p \int_{t=0}^{t_{exe}}\dddot{p}(t)^2dt + w_c J_c,\\
	& J_c = \left\{
	\begin{aligned}
		&(\mathcal{C} - c_{obs})^2 , &c_{obs} < \mathcal{C}, \\
		&0 , &c_{obs} \geq \mathcal{C},
	\end{aligned}
	\right.
\end{align}
where $\mathcal{C}$ is the desired clearance, the weights $w_t, w_p, w_c$ are set to 1, 0.01, and 100, respectively, so that three terms have comparable numerical magnitude. 
The problem formulation procedure is illustrated in Fig. \ref{pic:changing_env_method}.
When the current trajectory to execute is determined, the planner will select a point $P_s$ on the current trajectory under a given distance $R$ to the current start point immediately.
Then the parameter tuning procedure for the subsequent trajectory will start, and all the intermediate variables of previous optimizations will be cleared as the environment becomes different.
Before the drone gets close to $P_s$, many trajectories will be generated with different $\theta$, where the one with the smallest penalty value $f_s^*$ is selected.

In this experiment, the quadrotor moves with changing obstacle density, thus the parameters must converge quickly to adapt to the current environment.
Therefore like previous works \cite{quan2021eva,richter2014high}, we only optimize the desired velocity as well, which reduces the dimension of solution space.
As the convergence rate is highly concerned, and there will not generate too many trajectories in each round, Bayesian Optimization is adopted in short-term objective optimization in this experiment.

\begin{figure}[t]
	\centering
	\includegraphics[width=1.0\linewidth]{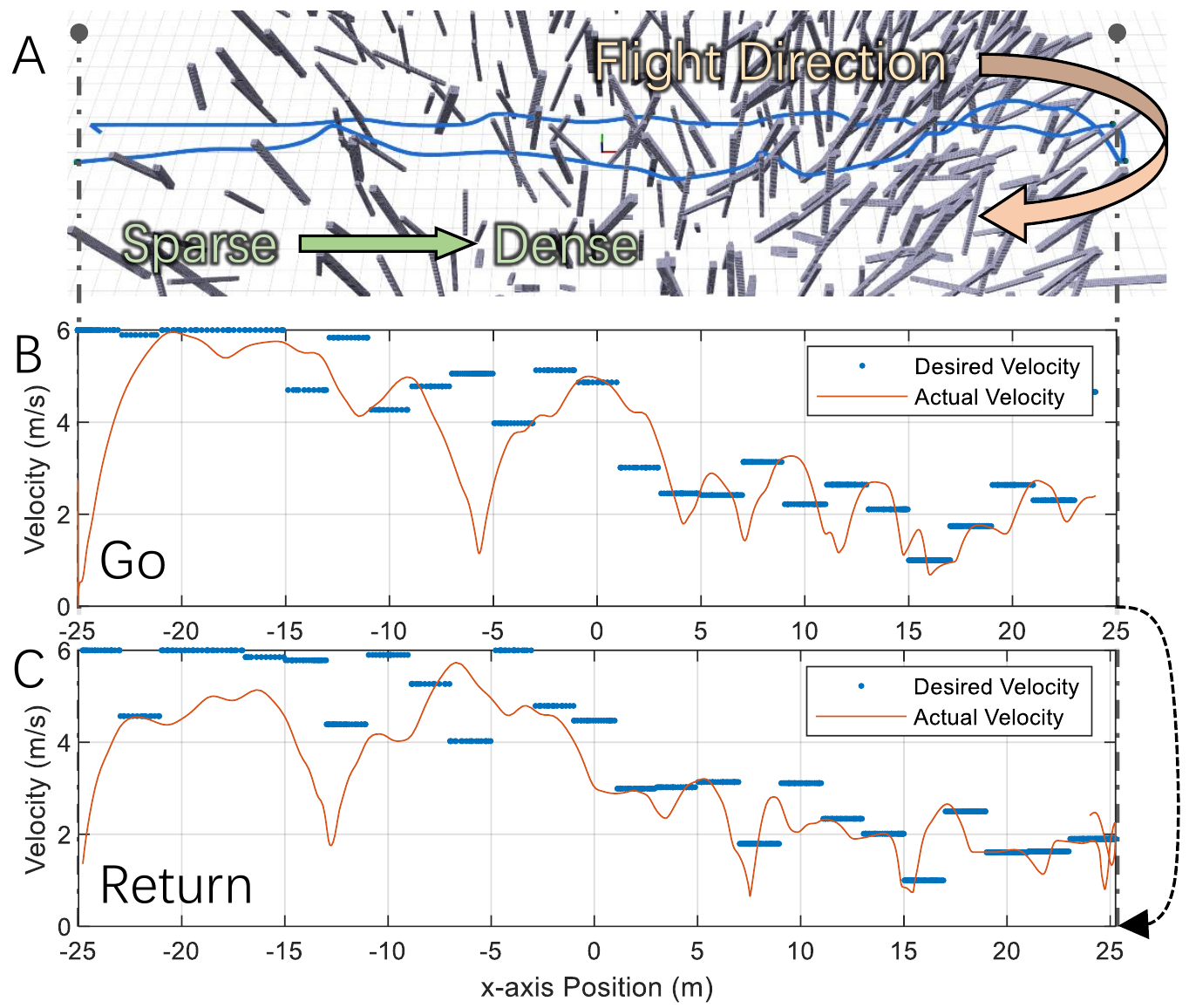}
	\caption{ A flight test in an environment with changing obstacle density. (A) A drone flies with changing obstacle distribution. (B, C) The desired flight speed changes in different regions.}
	\label{pic:chenging_env_result}	
\end{figure}

We conduct this experiment based on EGO-Planner \cite{zhou2020ego}. 
The result is shown in Fig. \ref{pic:chenging_env_result}.
The velocity boundary is set to 1 -- 6 m/s and the \textit{Desired Velocity} is the adaptive parameter.
Although we neither explicitly extract obstacle distributions nor specifically design speed altering rules like \cite{quan2021eva,richter2014high}, the flight speed still adapts to the environment.
Note that the actual velocity can not always catch up with the desired because of some inner mechanisms of the trajectory planner, which will make some manual velocity-tuning rules less reasonable.
The selection of $R$ balances optimality and reaction time.
A longer $R$ allows more computing time to try more parameters, while a shorter $R$ makes the parameter adapt to the environment more quickly.
Another adverse effect of long $R$ is that the environment observed by onboard sensors is also changing during the tuning procedure before the drone arrives at $P_s$, then early results may become less meaningful.
To address this problem, we gradually enlarge the observation noise over time of previously trialed trajectories.
Our experiments set $R$ to 2 m, which shows an acceptable balance between optimality and reaction time.

\subsubsection{Fixed Environments}

\begin{figure}[t]
	\centering
	\includegraphics[width=1.0\linewidth]{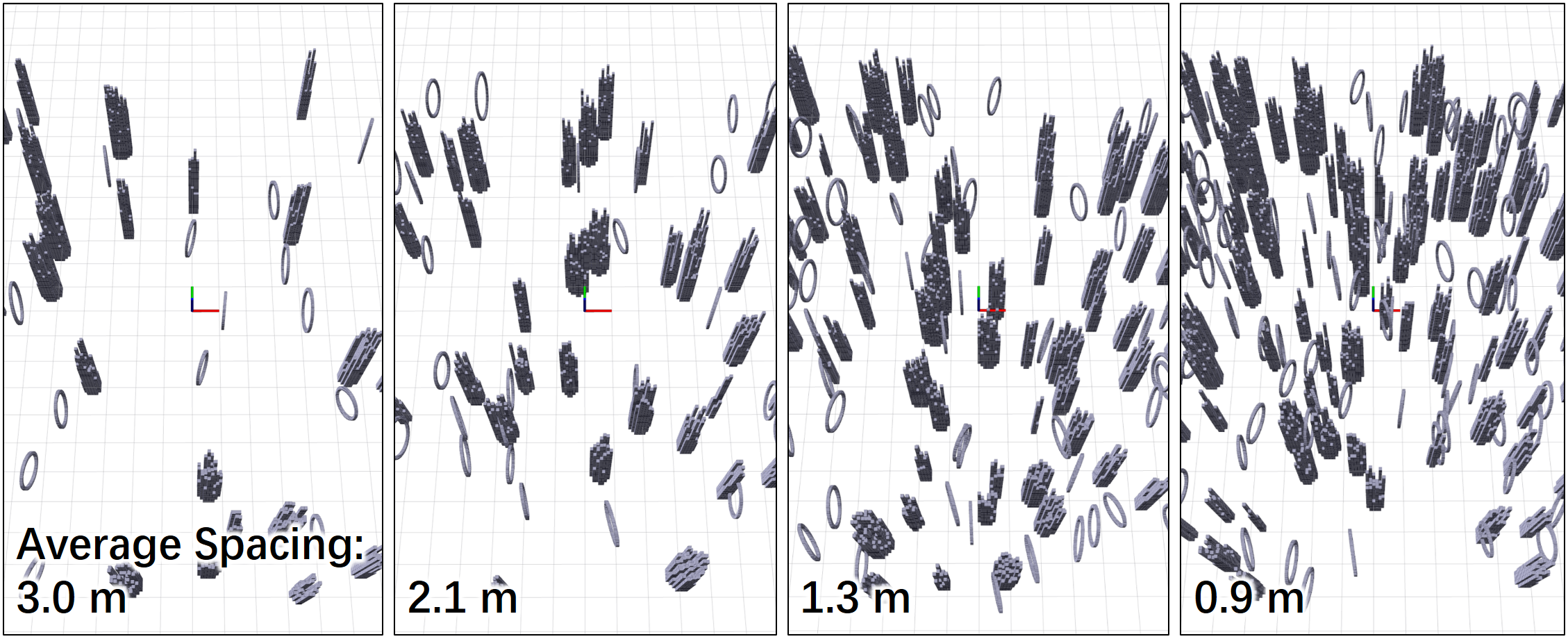}
	\caption{ Various obstacle densities. Obstacles are annular or cylindrical.}
	\label{pic:obstacle_distribution}	
		\vspace{-1.5cm}
\end{figure}

If the obstacle density is uniform throughout the entire map, as shown in Fig. \ref{pic:obstacle_distribution}, the objective definition and the optimized parameters can stay identical to Sec. \ref{sec:Benchmark_Comparisons}.
The drone is commanded to fly across the area with obstacles and then return to the start position.
The results shown in Table \ref{tab:different_env} are average values with a standard deviation of five repeats.
From the results, we can see that the proposed framework successfully functions in all four environments.

\begin{table}[]
	\renewcommand{\arraystretch}{1.35}
	\centering
	\caption{ Parameter tuning with various obstacle densities.}
	\begin{tabular}{|c|c|c|cc|}
		\hline
		\multirow{3}{*}{\textbf{\begin{tabular}[c]{@{}c@{}}Average \\ Spacing\\ (m)\end{tabular}}} & \multirow{3}{*}{\begin{tabular}[c]{@{}c@{}}Flight \\ Time\\ (s)\end{tabular}} & \multirow{3}{*}{\begin{tabular}[c]{@{}c@{}}Maximum\\ Emergency\\ Stops\end{tabular}} & \multicolumn{2}{c|}{\multirow{2}{*}{90\% Optimum Evaluations}} \\
		&                                                                         &                                                                                      & \multicolumn{2}{c|}{}                                          \\ \cline{4-5} 
		&                                                                         &                                                                                      & \multicolumn{1}{c|}{Short-Term}               & Long-Term            \\ \hline
		\textbf{3.0}                                                                                 & 11.25±0.85                                                              & 0                                                                                    & \multicolumn{1}{c|}{36215±4534}           & 7.6±3.1            \\ \hline
		\textbf{2.1}                                                                               & 12.56±1.23                                                              & 0                                                                                    & \multicolumn{1}{c|}{49235±12523}          & 15.2±4.2           \\ \hline
		\textbf{1.3}                                                                               & 19.53±2.02                                                              & 1                                                                                    & \multicolumn{1}{c|}{81354±21535}          & 15.6±5.5           \\ \hline
		\textbf{0.9}                                                                               & 41.8±8.44                                                               & 1                                                                                    & \multicolumn{1}{c|}{1.0e6±3.2e5}          & 45.2±12.2          \\ \hline
	\end{tabular}
\label{tab:different_env}
\end{table}

\subsection{Real-word Deployment with Different Quadrotors and Trajectory Planners}
\label{sec:Real_word_Deployment}

\begin{figure}[t]
	\centering
	\includegraphics[width=1.0\linewidth]{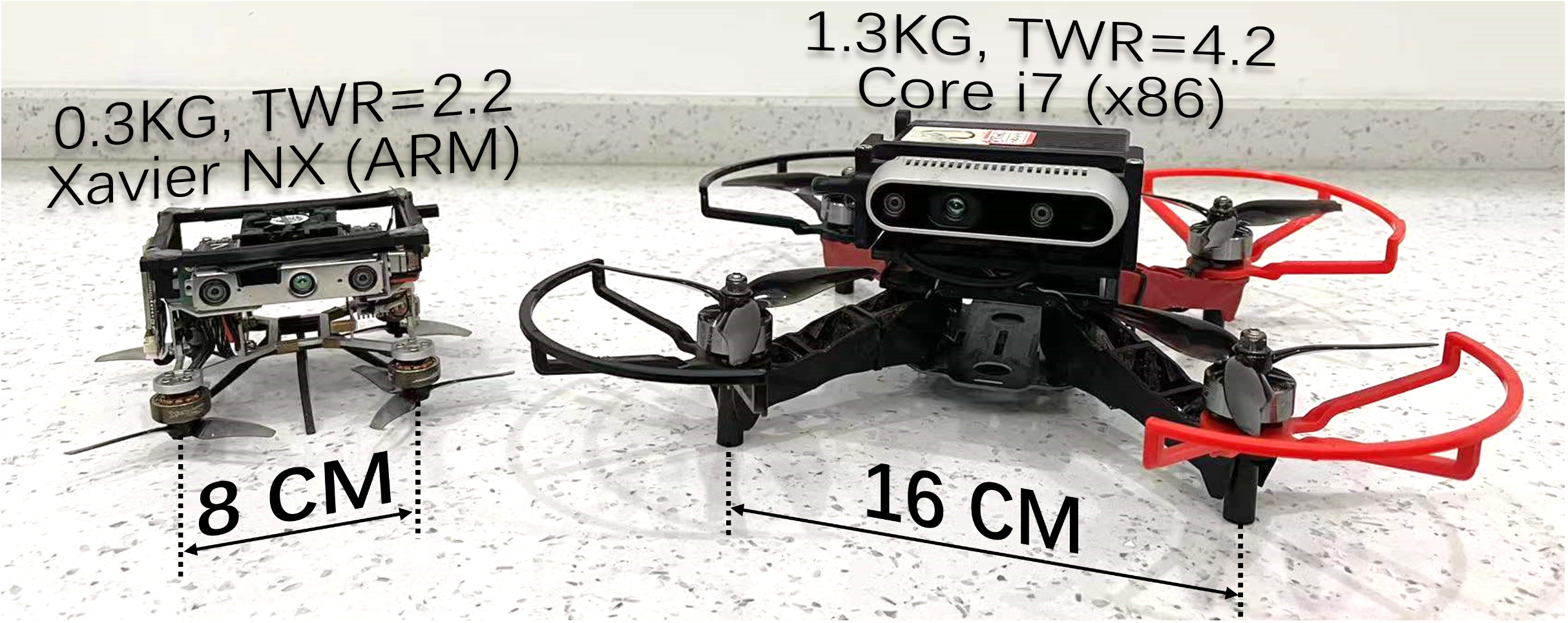}
	\caption{Two full-stack drone platforms used. \textit{TWR} indicates the thrust-to-weight ratio.}
	\label{pic:big_small_drone}	
		\vspace{-1.2cm}
\end{figure}

After getting a suitable parameter set from the simulation, we can move to real-world refinement and deployment.
To encourage more developers to incorporate our parameter tuning framework into their planners, we further validate its effectiveness on another online trajectory planning framework FAST-Planner \cite{zhou2019robust}.
It has a different collision avoidance formulation as well as a time adjustment mechanism.
Based on optimized parameters, following Sec. \ref{sec:sim_to_real}, we tested the proposed framework on two different drone platforms in different environments.
They differ in drone size, weight, onboard computers used (Fig. \ref{pic:big_small_drone}), and the obstacle density.

\subsubsection{Indoor, Big Drone, EGO-Planner}

\begin{figure}[ht]
	\centering
	\includegraphics[width=1.0\linewidth]{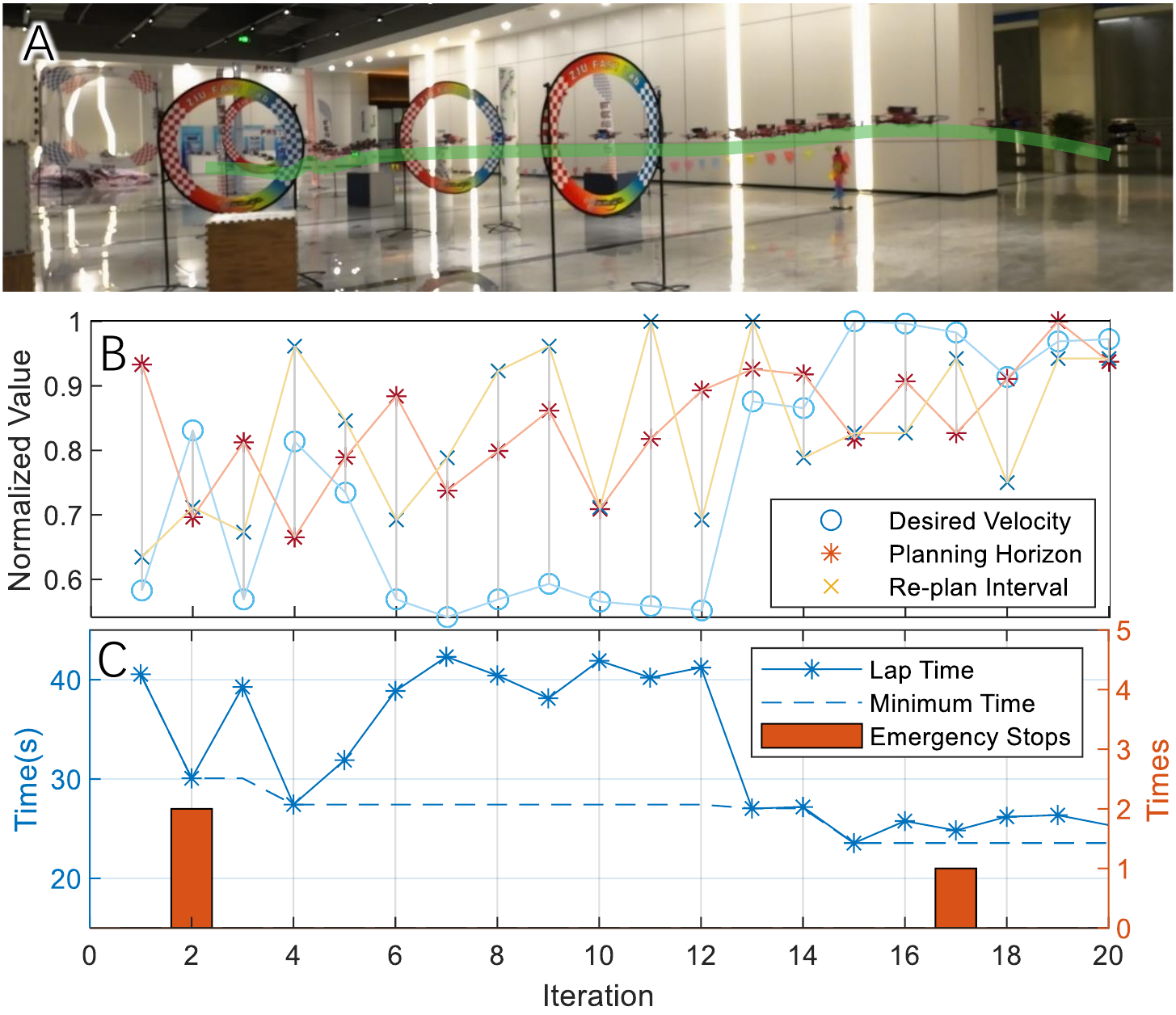}
	\caption{The indoor experiment using the big drone running EGO-Planner. (A) The testing environment. (B) Evolution of three long-term parameters during the optimization. (C) Flight performance. Note that C has two Y-axes.}
	\label{pic:indoor}	
\end{figure}

The big drone (the right one in Fig. \ref{pic:big_small_drone}) is used in the indoor experiment.
The optimized parameters are identical to Sec. \ref{sec:Benchmark_Comparisons}, and the drone also performs go-and-return behaviors for recording the flight time.
Values in Fig. \ref{pic:indoor}B are normalized for better visualization, where their coefficient to recover the initial value for \textit{Desired Velocity}, \textit{Planning Horizon}, and \textit{Re-plan Interval} are 2.90, 11.2, and 0.52, respectively.
From Fig. \ref{pic:indoor}B, we find that the optimizer falls into a local minimum between iteration 6 to 12 but managed to escape to another point with higher expected improvement. 
Finally the drone is able to fly across the obstacle-rich environment spending a shorter time than the initial parameters. 
Please refer to the attached video for an intuitive sense of the performance difference before and after optimization.
We validate the optimized parameters by extra five repeated flights; the flight time is 23.18±0.69 s with only one emergency stop occurred.

\subsubsection{Outdoor, Small Drone, FAST-Planner}

\begin{figure}[t]
	\centering
	\includegraphics[width=1.0\linewidth]{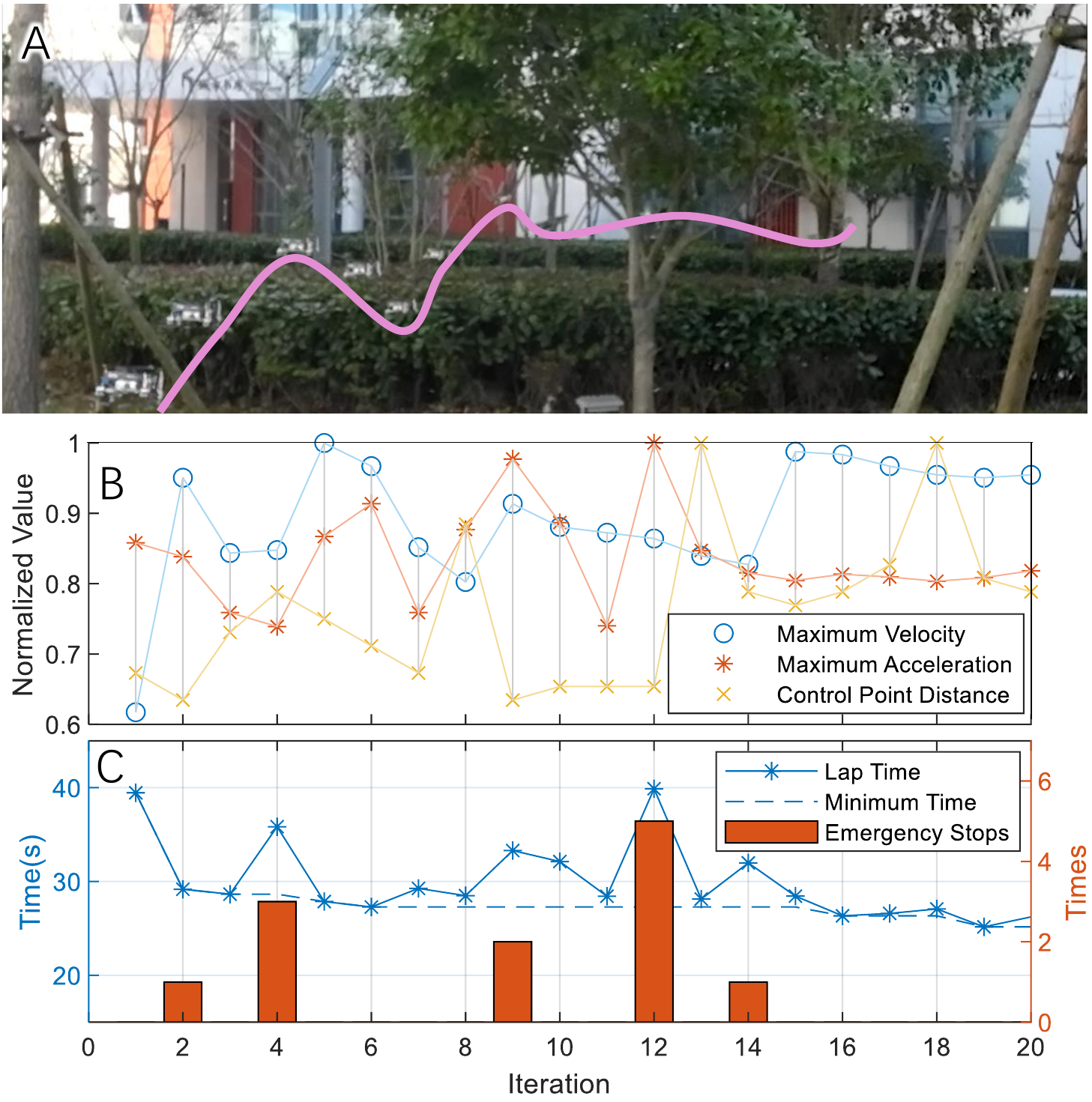}
	\caption{The outdoor experiment using the small drone running FAST-Planner. (A) The environment. (B) Evolution of three long-term parameters during the optimization. (C) Flight performance. Note that C has two Y-axes.}
	\label{pic:outdoor}	
		\vspace{-1.0cm}
\end{figure}

The small drone (the left one in Fig. \ref{pic:big_small_drone}) is used in the outdoor experiment.
FAST-Planner \cite{zhou2019robust} running on an NVIDIA Xavier NX onboard computer performs go-and-return flight as well.
Also, the flight time and the number of emergency stops are recorded.
On FAST-Planner, we optimize three parameters: the \textit{maximum velocity}, \textit{maximum acceleration}, and the \textit{control point distance}, whose corresponding coefficients are 2.40, 5.04, and 0.52, respectively.
From Fig. \ref{pic:outdoor}, Bayesian Optimization converges after the 15-th iteration and finally achieves short flight time and fewer emergency cases.
We also validate the final parameter for five flights and record a 26.53±1.21-s flight time with totally one emergency stop.
This outdoor optimization validates the effectiveness of adopting the proposed parameter adaption framework onto different situations.

\section{Conclusion and Future Work}

This paper proposes an automatic parameter adaption framework designed for online quadrotor trajectory planners.
It separates parameters into two groups that are optimized using different strategies.
This separation allows some parameters to get optimized hundreds of times per second while other objective functions can only be evaluated after a pried of fight.
The parameters optimized under a high frequency can further accelerate the convergence of the entire parameter adaption as both optimizers update parameters on the same planner $G$.
Simulation and real-world experiments validate the effectiveness of the proposed adaption framework in various environments using different planners.
Planner developers can incorporate this framework into their work to provide adaption in various use cases.

The proposed complementary parameter tuning framework also has the potential to be adapted to other systems.
For example, some parameters can be turned at a high frequency in controller tuning by predicting future control errors based on the system model, while others are optimized after observing the real actuating performance.
In visual-based localization, some objectives and parameters that are difficult to optimize, such as computing time, number of features to reserve, etc., can be optimized using the proposed framework by trial and error in multiple threads in real-time.

\newlength{\bibitemsep}\setlength{\bibitemsep}{.033\baselineskip}
\newlength{\bibparskip}\setlength{\bibparskip}{0pt}
\let\oldthebibliography\thebibliography
\renewcommand\thebibliography[1]{%
  \oldthebibliography{#1}%
  \setlength{\parskip}{\bibitemsep}%
  \setlength{\itemsep}{\bibparskip}%
}
\bibliography{iros2022Xin} 
\end{document}